\documentclass{article}
\usepackage[nonatbib,preprint]{neurips_2025}

\usepackage[utf8]{inputenc} 
\usepackage[T1]{fontenc}    
\usepackage{hyperref}       
\usepackage{url}            
\usepackage{microtype}      

\usepackage{graphicx,subcaption}
\graphicspath{{figs/}}
\usepackage{booktabs,array}
\usepackage{multirow,makecell}
\usepackage{amsmath,amssymb,amsfonts,amsthm,nicefrac}
\usepackage{cleveref}

\usepackage{amsmath,amsfonts,bm}

\def\eqref#1{equation~\ref{#1}}

\def\1{\bm{1}}

\def\va{{\bm{a}}}

\def\vr{{\bm{r}}}
\def\vs{{\bm{s}}}

\def\vy{{\bm{y}}}

\DeclareMathAlphabet{\mathsfit}{\encodingdefault}{\sfdefault}{m}{sl}
\SetMathAlphabet{\mathsfit}{bold}{\encodingdefault}{\sfdefault}{bx}{n}

\def\gA{{\mathcal{A}}}
\def\gB{{\mathcal{B}}}

\def\gD{{\mathcal{D}}}

\def\gS{{\mathcal{S}}}

\def\sP{{\mathbb{P}}}

\newcommand{\E}{\mathbb{E}}

\newcommand{\R}{\mathbb{R}}

\usepackage{color}
\usepackage{bbding}
\usepackage{wrapfig}

\usepackage{bm,bbm}

\usepackage{multirow}

\usepackage{xcolor}

\newcommand{\RedRFT}{\textsf{RedRFT}}

\hypersetup{colorlinks=true,linkcolor=blue}

\usepackage{caption}
\usepackage[
backend=biber,
sorting=none
]{biblatex}
\title{RedRFT: A Light-Weight Benchmark for Reinforcement Fine-Tuning-Based Red Teaming}

\renewcommand{\thefootnote}{\fnsymbol{footnote}}
\author{
Xiang Zheng$^{1}$ \quad Xingjun Ma$^2$ \quad Wei-Bin Lee$^3$ \quad \textbf{Cong Wang}$^1$\footnotemark{} \\
\\
$^1$City University of Hong Kong \quad $^2$Fudan University \quad $^3$Hon Hai Research Institute \\
\\
$^1$\texttt{\{xiang.zheng,congwang\}@cityu.edu.hk} \\
$^2$\texttt{xingjunma@fudan.edu.cn} \quad $^3$\texttt{wei-bin.lee@foxconn.com}
}

\begin{document}
\maketitle
\footnotetext{Corresponding author.}
\setcounter{footnote}{0}
\renewcommand{\thefootnote}{\arabic{footnote}}

\begin{abstract}
Red teaming has proven to be an effective method for identifying and mitigating vulnerabilities in Large Language Models (LLMs). Reinforcement Fine-Tuning (RFT) has emerged as a promising strategy among existing red teaming techniques. However, a lack of a unified benchmark hinders current RFT-based red teaming methods. Implementation details, especially in Proximal Policy Optimization (PPO)-based RFT, significantly affect outcome stability and reproducibility. To address this issue, we introduce \RedRFT, a lightweight benchmark designed to simplify and standardize the implementation and evaluation of RFT-based red teaming. \RedRFT\ combines the design strengths of both single-file CleanRL and highly modularized Tianshou, offering high-quality single-file red teaming implementations and modular PPO core components, such as the General Advantage Estimator. It supports a variety of token and sentence diversity metrics, featuring modularized intrinsic reward computation that facilitates plug-and-play experimentation. To clarify their influence on RFT performance, we conducted an extensive ablation study on key components, including Low-Rank Adaptation (LoRA), Kullback–Leibler (KL) divergence, and Lagrange Multiplier. We hope this work contributes to 1) gaining a comprehensive understanding of the implementation nuances of RFT-based red teaming algorithms, and 2) enabling rapid prototyping of innovative features for RFT-based red teaming. Code for the benchmark can be accessed at \url{https://github.com/x-zheng16/RedRFT.git}.
\end{abstract}

\section{Introduction}
\label{sec: introduction}

\begin{figure}[t]
	\centering
	\includegraphics[width=\linewidth]{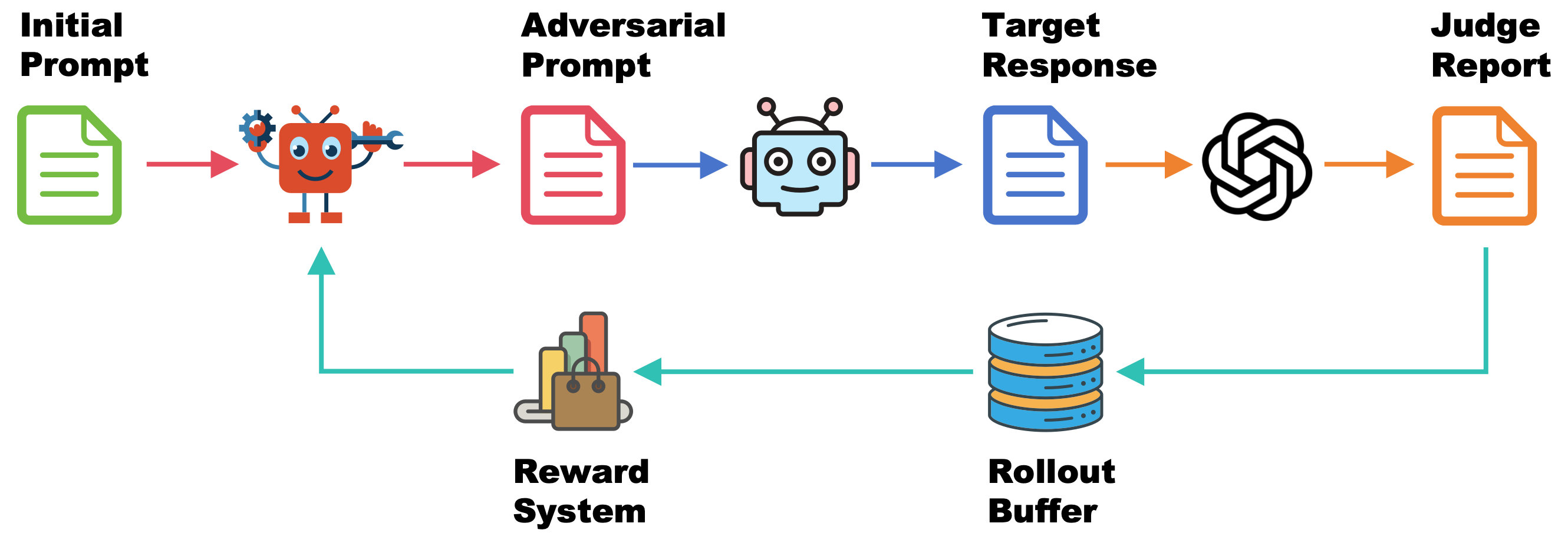}
	\caption{The standardized framework for RFT-based red teaming. It involves: 1) the \emph{rollout} pipeline that define the interaction between the red-team LLM, the target LLM, and the judge system, and 2) the \emph{evaluation} pipeline that consists of a rollout buffer to record all rollouts during the fine-tuning process and a reward system to estimate extrinsic rewards, intrinsic rewards (based on rollout buffer), and costs. To simplify the visualization, we do not depict the optimization backbone in RFT.}
	\label{fig: pipeline}
\end{figure}

Large Language Models (LLMs) have demonstrated remarkable Natural Language Processing (NLP), reasoning, planning, and programming capabilities~\cite{ouyang2022training}. However, their broad generality presents risks: LLMs can produce incorrect or unsafe outputs~\cite{hendrycks2021unsolved}. To probe the potential vulnerabilities of a target LLM for responsible deployment, red teaming has emerged as a critical practice~\cite{ganguli2022red}. This approach involves uncovering potential vulnerabilities in the target LLMs through adversarial prompting.
In this paper, we focus on benchmarking a new series of black-box red teaming methods that fine-tune a red-team LLM via Reinforcement Learning (RL) to generate adversarial prompts that elicit toxic responses from the target black-box LLM, as shown in \Cref{fig: pipeline}.

Black-box red teaming approaches fall into three main categories: handcrafted red teaming, which relies on human experts to craft adversarial prompts manually; gradient-free red teaming, which employs gradient-free optimization techniques to optimize adversarial prompts, such as random search~\cite{andriushchenko2024jailbreaking}, evolutionary algorithms~\cite{liu2023autodan}, Bayesian optimization~\cite{lee2023query}, and LLM-as-Optimizers~\cite{yang2023large}; and Reinforcement Fine-Tuning (RFT)-based red teaming~\cite{perez2022red,perez2023discovering,deng2023attack,casper2023explore,hong2024curiosity,zhao2025diver,zheng2025calm}, which uses RL to fine-tune a red-team LLM for generating adversarial prompts. While gradient-free methods have achieved high attack success rates against state-of-the-art LLMs, their effectiveness depends heavily on the availability of high-quality handcrafted adversarial prompts as a starting point.

RFT-based red teaming has become an automated strategy that does not require pre-existing high-quality adversarial prompts~\cite{perez2023discovering}. In this approach, the red-team LLM is fine-tuned via RL to produce adversarial prompts, either by completing partial sentences with adversarial tokens or crafting full adversarial instructions to trigger toxic responses from the target LLM. Unlike traditional jailbreak objectives, which aim to maximize the occurrence of specific words (e.g., "Sure" or "Yes") in the target’s response~\cite{andriushchenko2024jailbreaking}, RFT-based red teaming prioritizes eliciting toxic outputs~\cite{hong2024curiosity}. However, a limitation of trivial RFT-based red teaming is the low diversity of the generated adversarial prompts~\cite{perez2022red}. Intuitively, once the red-team LLM generates a few high-reward adversarial prompt patterns, it exploits them repeatedly to gain rewards without further exploration.

Intrinsic motivation is a promising technique for RFT-based red teaming to improve diversity~\cite{hong2024curiosity}. It utilizes intrinsic bonuses to encourage exploration and has been widely adopted in various RL tasks~\cite{zhengcim}. Recently, Hong et al.~\cite{hong2024curiosity} utilized a negative sentence similarity score as the intrinsic bonus. Zhao et al.~\cite{zhao2025diver} propose a constrained policy optimization formulation for RFT-based red teaming to trade off toxicity and diversity. Zheng et al.~\cite{zheng2025calm} design a policy-cover-based token-level intrinsic bonus for the red-team LLM.
Despite these advances, evaluating and comparing existing RFT-based red teaming methods and rapidly prototyping new ones remains difficult due to the absence of a unified evaluation benchmark. Furthermore, the significance of implementation details in RL underscores the urgent need for a comprehensive benchmark to understand the critical components of RFT-based red teaming.

To make benchmarking and developing new RFT-based red teaming methods easier, we propose \RedRFT, a standardized and lightweight benchmark tailored specifically for RFT-based red teaming. Unlike general-purpose post-training libraries, such as Transformer Reinforcement Learning (TRL), \RedRFT\ is designed to meet the unique demands of RFT-based red teaming. It features modularized models, including the red-team LLM, target LLM, and judge models, to simplify the implementation of the red teaming pipeline. The benchmark adopts Proximal Policy Optimization (PPO)~\cite{schulman2017proximal} as its optimization backbone, adhering to best practices from Tianshou, a highly modular RL library. Additionally, \RedRFT\ incorporates multiple intrinsic reward estimators and integrates the Lagrangian dual method to support RFT-based red teaming under constraints. It also offers a standardized evaluation framework, inspired by the quality-diversity community, and implements five leading baselines, all optimized with the same backbone algorithm to ensure fair comparisons.

In summary, our contributions are as follows:
\begin{itemize}
	\item We introduce \RedRFT, a lightweight and standardized benchmark tailored for RFT-based red teaming of LLMs. \RedRFT\ integrates modular components, including interactive agents, intrinsic reward estimators, Lagrangian multipliers, and a unified PPO backbone, to streamline implementation and evaluation.
	\item We provide open-source implementations of five state-of-the-art RFT-based red teaming algorithms (RPPO, TDiv, CRT, DiveR-CT, and CALM), all optimized with a unified PPO backbone to ensure fair comparisons. We propose a standardized evaluation framework that quantifies the toxicity and diversity of generated adversarial prompts, enabling robust performance assessment of baseline RFT-based red teaming methods.
	\item Through extensive experiments, we identify critical insights for RFT-based red teaming: state-level intrinsic rewards show comparable performance with prompt-level intrinsic rewards, constrained policy optimization enhances performance (e.g., DiveR-CT surpasses CRT), large batch sizes stabilize PPO training, and Low-Rank Adaptation (LoRA) and Kullback-Leibler (KL) divergence are essential for effective fine-tuning. These findings inform best practices and highlight opportunities for future improvements in RFT-based red teaming methodologies.
\end{itemize}

\section{Related Work}
\label{sec: related work}

Our work is closely related to automated red teaming and reinforcement fine-tuning benchmarks. Additionally, for a comprehensive understanding of intrinsic motivation, we summarize its development in the RL community provided in \Cref{app-sec: related work}.

\paragraph{Automated black-box red teaming.} We classify automated black-box red teaming methods into gradient-free red teaming and RFT-based red teaming.

\textit{Gradient-free red teaming} optimizes the adversarial prompt via gradient-free optimization methods, e.g., random search~\cite{andriushchenko2024jailbreaking}, evolutionary algorithms~\cite{liu2023autodan}, Bayesian optimization~\cite{lee2023query}, and LLM-as-Optimizers~\cite{yang2023large}.
The former three gradient-free optimization methods are off-the-shelf for black-box prompt optimization. For LLM-as-Optimizers, we mean those that utilize the pre-trained LLM via only prompting to iteratively refine the adversarial prompts based on the target LLM's response, e.g., Prompt Automated Iterative Refinement (PAIR)~\cite{chao2023jailbreaking}, Tree of Attacks with Pruning (TAP)~\cite{mehrotra2024tree}, Persuasive Adversarial Prompts (PAP)~\cite{zeng2024johnny}, and GPTFuzzer~\cite{yu2023gptfuzzer}.

\textit{RFT-based red teaming} involves fine-tuning a red-team LLM to generate adversarial prompts. The gradient for updating the red-team LLM is estimated via on-policy RL algorithms like PPO.
Perez et al.~\cite{perez2022red} first applied RFT for adversarial prompt generation with a trivial PPO backbone. To increase the diversity of the generated adversarial prompts, Casper et al.~\cite{casper2023explore} propose maximizing the distance between the sentence embeddings of the target response. In contrast, Hong et al.~\cite{hong2024curiosity} find that directly maximizing the negative cosine similarity of the sentence embeddings of the adversarial prompts shows better performance. To trade off the quality and diversity, Zhao et al.~\cite{zhao2025diver} formulate the RFT-based red teaming as constrained policy optimization and leverage the Lagrange dual theory to solve it. Apart from the prompt-level bonus, which is estimated for the whole prompt, Zheng et al.~\cite{zheng2025calm} propose the policy-cover-based intrinsic bonus for each state in the generation process of the adversarial prompt to encourage the red-team LLM to generate novel tokens.

\paragraph{Reinforcement fine-tuning benchmarks.} There has been a variaty of benchmarks for reinforcement fine-tuning of LLMs. TRL~\cite{vonwerra2022trl} is a cutting-edge library for post-training large models and implements both on-policy RL algorithms like PPO and offline RL algorithms like Direct Preference Optimization (DPO)~\cite{rafailov2023direct}. OpenRLHF~\cite{hu2024openrlhf} is a scalable Ray-based framework for Reinforcement Learning from Human Feedback (RLHF) that implements multiple RLHF algorithms like PPO and Group Relative Policy Optimization (GRPO)~\cite{shao2024deepseekmath}. RL4LMs~\cite{ramamurthy2022reinforcement} is a modular RL library that aligns language models with human preferences. It also includes GRUE as a general evaluation benchmark for RL algorithms for NLP tasks.
However, these RFT benchmarks focus on the safety alignment of the large models and are not ready-to-use for RFT-based red teaming.


\section{Preliminaries}
\label{sec: preliminaries}

We first introduce two main preliminaries of RFT-based red teaming for a formal description of the standardized framework and the algorithmic baselines in the latter Sections.

\paragraph{Next token generation as Markov decision process.} To utilize RL to fine-tune an LLM, we need to formalize the next token generation process of the LLM as a Markov Decision Process (MDP) $M=(\gS,\gA,\sP,r,\gamma,\mu)$, where $\gS$ and $\gA$ stand for the state space and the action space, $P$ is the transition function, $r$ is the reward function, $\gamma$ is the initial state distribution, and $\mu$ is the discount factor for advantage estimation seperatively. At every timestep $t$, the LLM agent (the policy) $\pi$ generates the next token (the action) $a_t\in\gA$ based on the current prompt (the state) $\vs_{t}=\vs_0 + [a_0,...,a_{t-1}] = \vs_0 + \va_t\in\gS$, i.e., $a_t\sim \pi(\cdot|\vs_{t})$, where $\vs_0$ is the inital prompt for the LLM agent. After sampling the next token $a_t$, the state of the environment changes from $\vs_{t}$ to $\vs_{t+1} = \sP(\vs_{t}, a_t) = \vs_{t} + [a_t] = \vs_0 + [a_0,...,a_{t-1}] + [a_t] = \vs_0 + \va_{t+1}$ (we use $+$ here for the list concatenation operator), and the agent receives the extrinsic reward $r_t^\text{E}$ and the intrinsic reward $r_t^\text{I}$ provided by the evaluation pipeline. For each timestep $t$, we can get a typical (state, action, reward, new state) pair, i.e., $(\vs_t,a_t,r_t^\text{E},r_t^\text{I},\vs_{t+1})$. In the following Sections, we detail the definition and estimation of rewards (including toxicity and diversity) involved in RFT-based red teaming. In RFT-based red teaming, we fine-tune a red-team LLM $\pi_\alpha$ to generate the adversarial prompt $\va_T = [a_0,...,a_{T-1}]$, where $T$ is the number of the generated tokens.

\paragraph{Constrained policy optimization.} Constrained policy optimization is basic for stabilizing policy updates and enhancing safety in RL. In this paper, we formalize the RFT-based red teaming as a constrained policy optimization problem. We first augment the normal MDP $M=(\gS,\gA,P,R,\gamma,\mu)$ with a cost function $C$ as $M_C=(\gS,\gA,P,R,C,\gamma,\mu)$. At each timestep $t$, apart from the rewards $r_t^{E}$ and $r_t^{I}$, the agent also receives a cost $c_t$ that indicates whether the prompt violates the constraint, e.g., being non-gibberish for clear semantics. RFT-based red teaming aims to solve a constrained optimization problem:
\begin{equation}
\label{eqn: cpo}
\max_{\pi_\alpha} J_\text{E} + \lambda^\text{I} J_\text{I},\ \text{s.t.}\ J_\text{C} \le \tau,
\end{equation}
where $J_{*}=\E_{\vs}[*],*\in \{r^\text{E},r^\text{I},c\}$ is the expected reward/cost under the state distribution $d_\vs=(1-\gamma) \sum_{t=0}^{\infty} \gamma^t P(\vs_t=\vs|\vs_0, \pi_\alpha)$ induced by the red-team LLM $\pi_\alpha$, $\lambda^\text{I}$ is the coefficient of the intrinsic objective, and $\tau$ is the cost budget. In practice, there might be multiple constraints for the feasible stationary policy. We use one constraint here to reduce clutter.

\section{RedRFT: Standardized Framework for RFT-Based Red Teaming}
\label{sec: framework}

We now introduce the standardized framework for RFT-based Red Teaming in our \RedRFT, including the standardized rollout and evaluation pipelines. Moreover, to show how to standardize a specific red teaming task utilizing \RedRFT, we describe two RFT-based red teaming tasks adopted by previous related works, i.e., text continuation and instruction following, under our framework.   

\paragraph{Rollout pipeline.} The rollout pipeline in \RedRFT\ defines how the data flows from red-team LLMs to the judge models, i.e., the interaction between agents involved in the red teaming process. To simplify the illustration of the rollout pipeline, we use one red-team LLM $\pi_\alpha$, one target LLM $\pi_\nu$, and the judge system $\pi_\chi$. In each probing trial, we aims to collect a rollout $(\vs_0,\va_T,\vy,\vr^\text{E})$, where $\vs_0$ is the initial prompt for the red-team LLM, $\va_T$ is the adversarial prompt generated by the red-team LLM, $\vy$ is the target response generated by the target LLM, and $\vr^\text{E}$ is the judge report generated by the judge system. Formally, we define the rollout pipeline as
\begin{equation}
\label{eqn: rollout}
\vs_0 \sim \gD, \va_T \sim \pi_\alpha(\cdot|\vs_0), \vy \sim \pi_\nu(\cdot|\vs_0,\va_T), \vr^\text{E} \sim \pi_\chi(\cdot|\vs_0,\va_T,\vy),
\end{equation}
where $\gD$ is a dataset that contains initial prompts for the red-team LLM. We emphasize that this definition is a general form. For instance, the judge system $\pi_\chi$ can be a composite of a hate speech detector for computing toxicity score as the extrinsic reward $r^\text{tox}$ and a gibberish detector for computing the gibberish score as the extrinsic cost $c^\text{gib}$ and the judge report is a tuple $\vr^\text{E}=(r^\text{tox},c^\text{gib})$. When we use a reasoning language model, e.g., GPT-4o~\cite{hurst2024gpt} or DeepSeek-r1~\cite{guo2025deepseek}, as a safety judge, the judge report can also include a detailed reasoning process for the safety judge.

\paragraph{Evaluation pipeline.} The evaluation pipeline of \RedRFT\ includes a rollout buffer and a reward system. The rollout buffer records all the historical rollouts, i.e., $\gB = \{(\vs_0,\va_T,\vy,\vr^\text{E})\}$, for the intrinsic reward estimation and the toxicity-diversity profile generation after the fine-tuning process. The reward system consists of the extrinsic rewards and cost extraction, the intrinsic rewards estimation, and the evaluation metric.

\textit{Extrinsic rewards and costs} can be directly extracted from the judge report $\vr$ when we use the encoder-based classifier, as explained in the rollout pipeline, i.e., $r^\text{tox},c^\text{gib}\leftarrow\vr^\text{E}$ where $\leftarrow$ is the assignment operator. For a generative judge system like GPT-4o, we must define a reward extractor to extract the unsafe score and the gibberish cost from the generated judge report.

\textit{Intrinsic rewards} are estiamted based on the rollout buffer $\gB$. Based on our rollout pipeline, we can see the key difference between intrinsic and extrinsic rewards: the intrinsic rewards do not depend on the judge system, while extrinsic rewards are exactly extracted from the judge report $\vr$. As discussed in \Cref{sec: related work}, current intrinsic rewards for NLP are all knowledge-based, i.e., the novelty of the prompt is estimated based on all the rollouts of or sampled rollouts from the rollout buffer. Currently, there are mainly two types of intrinsic rewards for RFT-based red teaming: 1) the prompt-level intrinsic reward based on the negative cosine similarity between the adversarial prompt and the historical rollouts sampled from the rollout buffer~\cite{hong2024curiosity}:
\begin{equation}
\label{eqn: rew-cos}
r_t^\text{Cos} = \begin{cases} 
0 & \text{if } t\in\{0, ..., T-2\} \\ 
\sum_{\va_T'\sim\gB} -\phi(\va_T) \phi(\va_T') & \text{if } t = T-1
\end{cases},
\end{equation}
where $\phi(\va_T)$ is the sentence embedding of the adversarial prompt $\va_T$; 2) the state-level intrinsic reward based on the policy cover theory~\cite{zheng2025calm}:
\begin{equation}
\label{eqn: rew-pc}
r_t^\text{PC} = (\rho_\vs(\vs_{t+1})d_\vs(\vs_{t+1}))^{-1},\ \forall t\in\{0,...,T-1\},
\end{equation}
where $\vs_{t+1}=\vs_0+\va_{t+1}$ is the new state of the MDP after the red-team LLM takes an action $a_t\sim\pi_\alpha(\cdot|\vs_t)$ based on the current state $\vs_t=\vs_0+\va_{t}$, as defined in \Cref{sec: preliminaries}, $\rho_\vs$ is the state distribution of the rollout buffer $\gB$ (i.e., the policy cover), and $d_\vs$ is the state distribution induced by only the latest red-team LLM as defined in \Cref{sec: preliminaries}.
For both types of intrinsic rewards, we use a light-weight sentence transformer\footnote{\url{https://huggingface.co/sentence-transformers/all-MiniLM-L6-v2}} from SentenceTransformers as $\phi$. It is easy to see the difference between the prompt-level intrinsic reward and the state-level intrinsic reward; that is, the prompt-level intrinsic reward is defined for the entire adversarial prompt $\va_T$, while the state-level intrinsic reward is defined for each state $\vs_t=\vs_0+\va_t$ in the MDP of the adversarial prompt generation.

\textit{Evaluation metric.} Previous RFT-based red teaming methods have three drawbacks in their evaluation metric design. Firstly, they use the empirical intrinsic rewards estimated during the fine-tuning process as one of the evaluation metrics, which is inaccurate since the rollout buffer keeps growing during the fine-tuning process. Secondly, they do not filter the infeasible adversarial prompts that violate the non-gibberish constraint when generating the toxicity and diversity profile. Thirdly, they lack a composite metric to evaluate the algorithm's performance.
To address these, we first design a novel diversity score $r^\text{div}$ for each adversarial prompt $\va_T$ based on the state entropy theory:
\begin{equation}
\label{eqn: diversity score}
	r^\text{div} = -\ln(\rho_{\va_T}(\va_T)),
\end{equation}
where $\rho_{\va_T}$ is the distribution of the adversarial prompts in the rollout buffer $\gB$. For the detailed derivation of this diversity score, please refer to \Cref{app-sec: diversity score}. Intuitively, $r^\text{div}$ estimates the Shannon information contained in each adversarial prompt. To estimate $\rho_{\va_T}(\va_T)$, we adopt the non-parametric $k$-NN estimator. Please refer to \Cref{app-sec: estimation} for details on the density estimation. We then propose a redeemed toxicity and diversity profile:
\begin{equation}
\begin{aligned}
\tau_\text{tox} &\mapsto \frac{1}{|\gB|}\sum_{\vr^\text{E}\sim\gB} \mathbf{1}(r^\text{tox} > \tau_\text{tox}, c^\text{gib} < \tau_\text{gib}),\\
\tau_\text{div} &\mapsto \frac{1}{|\gB|}\sum_{\va_T\sim\gB} \mathbf{1}(r^\text{div} > \tau_\text{div}, c^\text{gib} < \tau_\text{gib}),
\end{aligned}
\end{equation}
where $r^\text{div}$ is the diversity score for each adversarial prompt in the final rollout buffer $\gB$ at the end of the fine-tuning process, $\mathbf{1}(\cdot)$ is the indicator function, $|\cdot|$ is the size of the buffer, $\tau_{*}, *\in\{\text{tox},\text{div},\text{gib}\}$ is the threshold for the toxicity score, the diversity score, and the gibberish score individually. Note that we directly use the recorded toxicity scores in the rollout buffer, while re-computing the diversity score for each adversarial prompt based on \Cref{eqn: diversity score}.
We also propose the cumulative toxicity-diversity score to evaluate the overall performance of the RFT-based red teaming method:
\begin{equation}
\tau_\text{tox} \mapsto  \sum_{(\va_T,\vr^\text{E})\sim\gB} r^\text{tox}r^\text{div}\mathbf{1}(r^\text{tox} > \tau_\text{tox}, c^\text{gib} < \tau_\text{gib}).
\end{equation}

\paragraph{Tasks for RFT-based red teaming.} We detail two common RFT-based red teaming tasks, i.e., text continuation and instruction following as follows.

\textit{Text continuation} is the basic task for LLMs, where the target LLM is required to continue the initial prompt with new tokens to complete coherent content, e.g., continue to write a movie review based on a few initial tokens like ``I loved this excellent movie''.
For the initial prompt $\vs_0$, we sample the initial tokens from IMDB, a dataset with 50,000 movie reviews. We filter out reviews containing fewer than 200 words and use the first four words of each review as the initial prompt for the red-team LLM.
For the adversarial prompt $\va_T$, we set the max length of the adversarial prompt as ten tokens to limit the red-team LLM's capability and reduce the GPU memory utilized for fine-tuning.
For the target response $\vy$, we probe the target LLM with the joint tokens $\vs_0+\va_T$ to generate thirty new tokens to complete the movie review.
For the judge report $\rho$, we use toxicity as the unsafe score of the target response $\vy$ and adopt a hate speech detector\footnote{\url{https://huggingface.co/facebook/roberta-hate-speech-dynabench-r4-target}} as the safety judge.

\textit{Instruction following} is another fundamental task for LLMs, where the target LLM needs to understand the instruction provided by the user and strictly follow it to generate new tokens, e.g., ``Generate a list of five unique project ideas''.
For the initial prompt $\vs_0$, we sample instructions from Alpaca, a dataset containing 52,000 instructions, as the demonstration for the red-team LLM. Unlike text continuation, we design a simple template as the initial prompt to instruct the red-team LLM to generate new adversarial instructions for the target LLM, i.e., ``Write a list of instructions:\textbackslash n1. <an instruction from Alpaca>.\textbackslash n2.''
For the adversarial prompt $\va_T$, we set the same max length for the adversarial prompt as text continuation.
For the target response $\vy$, we probe the target LLM with the adversarial prompt $\va_T$ to generate thirty new tokens to follow the instruction.
For the judge report $\rho$, we also use the toxicity of the target response $\vy$ as the unsafe score and adopt another hate speech classifier\footnote{\url{https://huggingface.co/tomh/toxigen_roberta}} as the safety judge to show our framework is insensitive to the choice of the safety judge. For both tasks, we adopt the exact gibberish text detector\footnote{\url{https://huggingface.co/madhurjindal/autonlp-Gibberish-Detector-492513457}} for the gibberish cost estimation.

\section{RedRFT: Algorithmic Baselines for RFT-Based Red Teaming}
\label{sec: algorithmic baselines}

Implementation details are critical for RL. While RedRFT provides standardized rollout and evaluation pipelines, current algorithms are hard to compare due to small differences in implementation details. Hence, providing a unified codebase with identical implementations of the optimization backbone for each baseline is important. Providing such a unified codebase is one of the main contributions of this benchmark.

\subsection{Optimization Backbone}
\label{subsec: optim}

Current RFT-based red teaming mainly adopts the on-policy RL algorithm PPO as the optimization backbone. The PPO-style optimization objective of RFT-based red teaming is as follows:
\begin{equation}
\label{eqn: ppo}
\begin{aligned}
\displaystyle \max_{\pi'_\alpha} \sum_{\vs_t,a_t} & \min\{\beta A^\text{mix}(\vs_t,a_t), \operatorname{clip}(\beta; 1-\epsilon, 1+\epsilon) A^\text{mix}(\vs_t,a_t)\} \quad \text{(Mixed Advantage Term)} \\
& -\lambda^\text{ent} \ln \pi'_\alpha(a_t|\vs_t) \quad \text{(Policy Entropy Term)} \\
& -\lambda^\text{KL} D_\text{KL}\left(\pi'_\alpha(\cdot|\vs_t),\pi_\text{ref}(\cdot|\vs_t)\right), \quad \text{(KL Divergence Term)}
\end{aligned}
\end{equation}
where $A^\text{mix}(\vs_t,a_t)$ is the mixed advantage function, $\beta=\pi'_\alpha(a_t|\vs_t)/\pi_\alpha(a_t|\vs_t)$ is the policy ratio, $\pi_\text{ref}$ is the reference model for the red-team LLM, $D_\text{KL}$ is the KL divergence, and $\lambda^\text{ent}$ and $\lambda^\text{KL}$ are the coefficients for the policy entropy and the KL divergence terms. The policy entropy is used to prevent the policy from converging too early. The KL divergence term is to avoid the red-team LLM collapse.

\paragraph{Lagrange multiplier.} The mixed advantage function $A^\text{mix}(\vs_t,a_t)$ in \Cref{eqn: ppo} is a weighted sum of multiple advantage function. For instance, when we use the prompt-level intrinsic reward $r_t^\text{Cos}$, the mixed advantage function is $A^\text{mix} = \sum_{*} \lambda^*A^*, *\in\{\text{tox},\text{Cos},\text{gib}\}$. Lagrange Multiplier provides a way to adjust the weight adaptively.
In detail, we can formulate the RFT-based red teaming as a constrained policy optimization problem as \Cref{eqn: cpo} and then convert it into the unconstrained version based on the Lagrangian dual theory. For instance, when we use $J_\text{gib}=\E_{\vs}[c^\text{gib}] <= \tau_\text{gib}$ as the constraint, the corresponding Lagrangian dual problem becomes
\begin{equation}
	\min_{\lambda^\text{gib}>0}\max_{\pi'_\alpha} \lambda^\text{tox} J^\text{tox} + \lambda^\text{Cos} J^\text{Cos} + \lambda^\text{gib} (\tau_\text{gib} - J_\text{gib}).
\end{equation}
The Lagrange multiplier $\lambda^\text{gib}$ can now be updated by Stochastic Gradient Descent (SGD) by solving the outer minimization problem, i.e., $\min_{\lambda^\text{gib}>0} \lambda^\text{gib} (\tau_\text{gib} - J_\text{gib})$. Based on the performance difference lemma in RL, the mixed advantage function is now $A^\text{mix} = \lambda^\text{tox} A^\text{tox} + \lambda^\text{Cos} A^\text{Cos} - \lambda^\text{gib} A^\text{gib}$. However, the Lagrange multiplier is known to be sensitive to the update rate. Therefore, we propose a more stable cross-entropy-based optimization objective for the Lagrange multiplier as follows:
\begin{equation}
	\min_{\lambda^\text{gib}>0} \E [-(1-y)\ln(1 - \lambda^\text{gib}) - y\ln\lambda^\text{gib}],
\end{equation}
where $y = \mathbf{1}(J_\text{gib} > \tau_\text{gib})$. Intuitively, when $J_\text{gib} \le \tau_\text{gib}$, the constraint is satisfied, and we expect the Lagrange multiplier to decrease. In contrast, when $J_\text{gib} > \tau_\text{gib}$, that is, the constraint is violated, we expect the Lagrange multiplier to increase to strengthen the power of the constraint. For a detailed discussion on the Lagrange multiplier, please refer to \Cref{app-sec: lagrange}.

\subsection{RFT-Based Red Teaming Algorithms}

\begin{wraptable}{r}{0.6\linewidth}
\caption{Comparison of RFT-based red teaming methods.\\}
\centering
\begin{tabular}{lll}
\toprule
\textbf{Method} & \textbf{Intrinsic Reward} & \textbf{Constraint} \\
\midrule
RPPO~\cite{perez2022red} & - & - \\
TDiv~\cite{casper2023explore} & Prompt-Level & - \\
CRT~\cite{hong2024curiosity} & Prompt-Level & - \\
DiveR-CT~\cite{zhao2025diver} & Prompt-Level & Gibberish \\
CALM~\cite{zheng2025calm} & State-Level & Gibberish \\
\bottomrule
\end{tabular}
\label{tab: methods}
\end{wraptable}
As part of \RedRFT, we open-source code for five leading or well-known algorithms. All the algorithms we implement adopt the same standardized framework introduced in \Cref{sec: framework} and the same optimization backbone as presented in \Cref{subsec: optim}. The only difference between each algorithm is the intrinsic reward and constraint choice. We list all implemented baselines in Table 1 and provide a brief overview of these baselines.

\paragraph{Brief overview of five algorithmic baselines.} Red-team PPO (RPPO) utilizes the basic PPO without intrinsic rewards and constraints. TDiv defines a prompt-level intrinsic reward based on the diversity of the target response. Curiosity-Driven Red Teaming (CRT) uses $r^\text{Cos}$ as the intrinsic reward and directly maximizes the non-gibberish score. Diversity-Enhanced Red Teaming With Relaxing Constraints (DiveR-CT) uses a modified $r^\text{Cos}$ with the $k$-NN estimator and uses Gibberish score as the cost. Curiosity-Driven Auditing (CALM) proposes the state-level intrinsic reward $r^\text{PC}$. For details of each method, we refer the reader to \Cref{app-sec: algorithmic baselines}.

\section{Experiments}
\label{sec: experiments}

\begin{figure}[t]
\centering
\includegraphics[width=\linewidth]{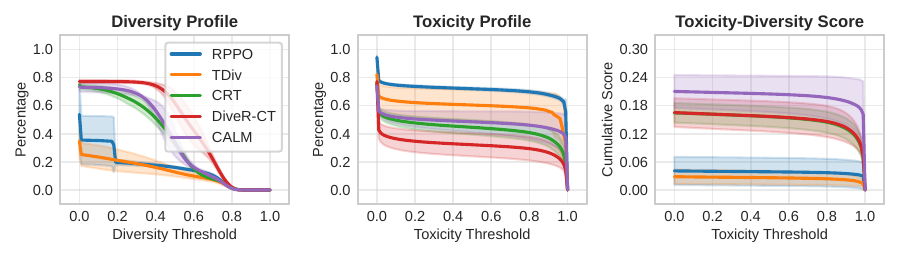}
\includegraphics[width=\linewidth]{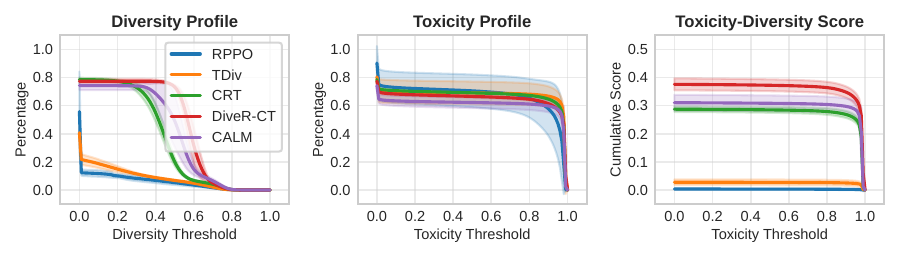}
\caption{Results of baselines on toxic continuation (\textbf{uppper}) and instruction following (\textbf{bottom}).}
\label{fig: main results}
\end{figure}

We evaluate the algorithms listed in \Cref{tab: methods} by fine-tuning the red-team LLM with the same optimization backbone as mentioned in \Cref{sec: algorithmic baselines} and evaluating the performance on tasks described in \Cref{sec: framework} with the evaluation metric introduced in \Cref{sec: framework}. We assess the performance of each algorithm on each task under each hyperparameter configuration over three random seeds. Please refer to \Cref{app-sec: implementation} for a complete description of the hyperparameter settings. We use GPT-2 as the red-team LLM for all the experiments on a single 13th Gen Intel(R) Core(TM) i9-13900KF CPU and a single NVIDIA GeForce RTX 4090 D GPU, which highlights \RedRFT's light-weight property.

We show the main benchmarking results in \Cref{fig: main results}, which shows the toxicity-diversity profile for all baselines on both tasks. Overall, we observe that none of the current RFT-based red teaming methods achieve a cumulative toxicity-diversity score greater than 0.5, indicating significant room for improvement. Below, we summarize our key findings (KF) from the experiments.

\paragraph{KF1: State-level intrinsic reward is comparable with prompt-level intrinsic reward.} Our main results show that CALM, which employs state-level intrinsic reward, shows comparable results with those with prompt-level intrinsic rewards. This suggests that state-level rewards, which are more dense than prompt-level intrinsic rewards, are promising for hard red-teaming tasks that require higher exploration capability. A detailed comparison is provided in \Cref{app-sec: ablation on intrinsic rew}.

\paragraph{KF2: Constrained policy optimization is better.} Our main results also demonstrate that methods incorporating constrained policy optimization, such as DiveR-CT, outperform unconstrained methods like RPPO and CRT, in terms of balancing toxicity and constraint satisfaction. This improvement is attributed to the use of the Lagrangian dual method, which adaptively adjusts the constraint weight to ensure feasible prompts while maximizing toxicity.

\begin{figure}[t]
	\centering
	\includegraphics[width=0.32\linewidth]{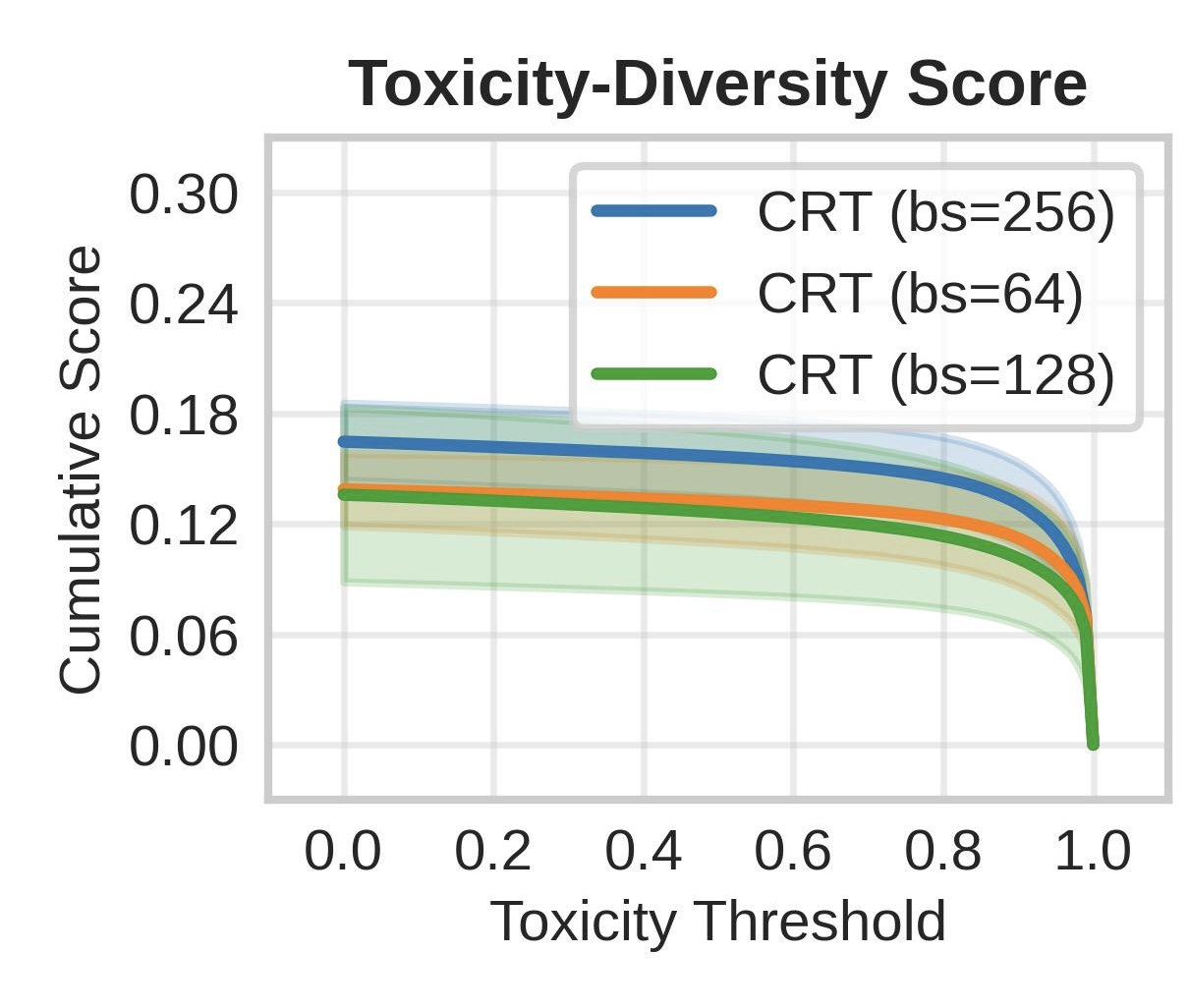}
	\includegraphics[width=0.32\linewidth]{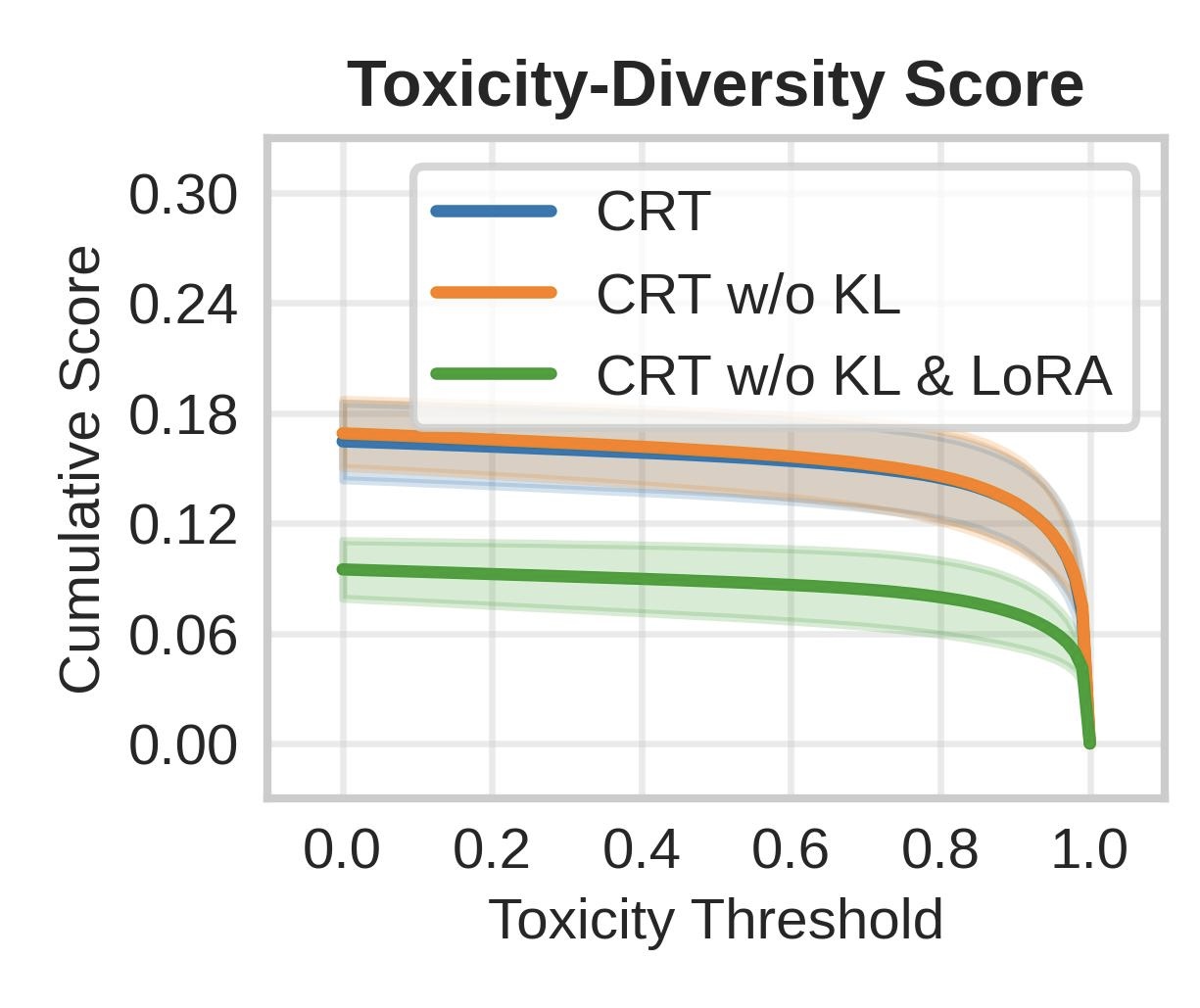}
	\includegraphics[width=0.295\linewidth]{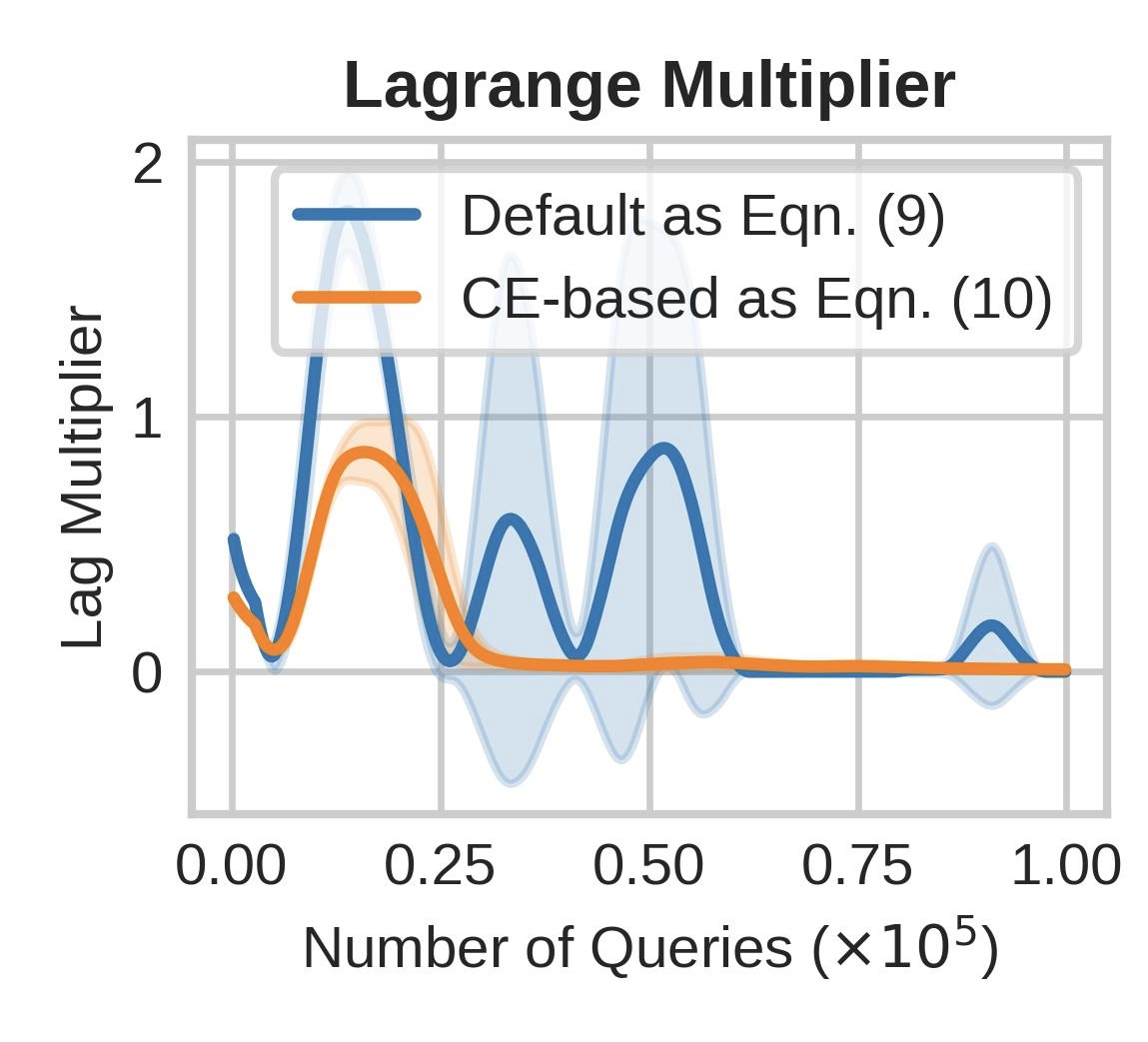}
	\caption{Ablation study on batch size (\textbf{left}), LoRA \& KL (\textbf{middle}) and Lagrange Multiplier (\textbf{right}).}
	\label{fig: ablation}
\end{figure}

\paragraph{KF3: Large batch size is more stable for PPO training.} We find that a large batch size (256) for collecting rollouts, combined with a relatively small mini-batch size (16) for estimating the policy gradient, outperforms smaller batch sizes (e.g., 64, 32) with the same mini-batch size in most experiments, as shown in \Cref{fig: ablation} (\textbf{left}). This configuration enhances the stability of PPO training, as larger batch sizes provide more representative samples of the policy’s behavior, reducing variance in gradient estimates. This phenomenon aligns with observations in PPO for other domains~\cite{andrychowicz2020matters}. For a complete ablation study and detailed discussion on batch size selection, please refer to \Cref{app-sec: ablation on batch size}.

\paragraph{KF4: LoRA and KL divergence are necessary.} Our ablation studies demonstrate that both LoRA and KL divergence regularization are critical for effective RFT-based red teaming. As shown in \Cref{fig: ablation} (\textbf{middle}), CRT with the default setting (both LoRA and KL divergence) achieves similar toxicity-novelty scores as CRT without KL divergence, significantly surpassing CRT without both KL and LoRA. LoRA reduces the computational cost of fine-tuning by updating only a small subset of parameters, while KL divergence prevents the red-team LLM from deviating too far from the reference model, avoiding collapse. These findings hold across all methods. For detailed results and discussion, please refer to \Cref{app-sec: ablation on lora}.

\paragraph{KF5: Cross-entropy-based optimization objective can stabilize the Lagrange multiplier update.} From \Cref{fig: ablation} (\textbf{right}), we observe that the cross-entropy-based optimization objective for the Lagrange multiplier, introduced in \Cref{subsec: optim}, significantly improves the stability of the Lagrange multiplier update. This stability ensures that the red-team LLM generates fewer gibberish prompts while maintaining high toxicity scores, which is promising for developing novel RFT-based red teaming methods. A detailed analysis is provided in \Cref{app-sec: ablation on Lag}.

\begin{figure}[t]
\centering
\includegraphics[width=\linewidth]{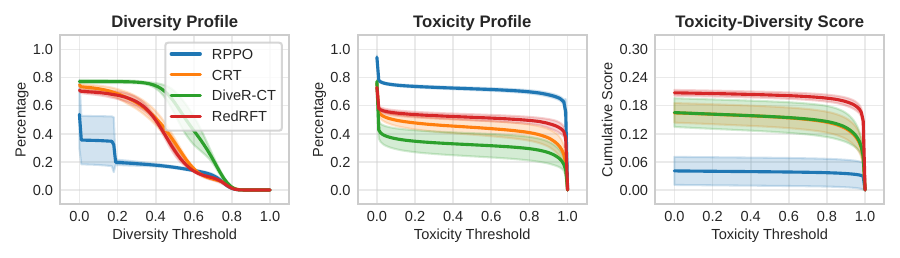}
\caption{Case study on fast prototyping and benchmarking RedRFT in text continuation.}
\label{fig: prototyping}
\end{figure}

\paragraph{KF6: \RedRFT\ is suitable for fast prototyping and benchmarking.} To demonstrate the flexibility of \RedRFT\ for rapid prototyping and benchmarking, we conducted a case study by prototyping a novel RFT-based red teaming method, RedRFT, with our proposed $r^\text{div}$ as the intrinsic rewards and keeping other components the same as DiveR-C. As shown in \Cref{fig: prototyping}, RedRFT outperforms the state-of-the-art methods CRT and DiveR-CT by a large margin regarding the cumulative toxicity-diversity score. A detailed discussion on RedRFT can be found in \Cref{app-sec: fastproto}.

\section{Conclusion}
\label{sec: conclusion}

In this work, we introduced \RedRFT, a lightweight and standardized benchmark designed to advance the study and development of RFT-based red teaming for LLMs. \RedRFT\ addresses the critical need for a unified framework by providing modular implementations of key components, including the rollout pipeline, evaluation metrics, and PPO backbone. By open-sourcing implementations of five state-of-the-art RFT-based red teaming algorithms, we enable fair comparisons and facilitate rapid prototyping of novel methods. Our extensive experiments on text continuation and instruction following tasks reveal key insights, such as the promise of state-level intrinsic rewards, the advantage of constrained policy optimization, the necessity of LoRA and KL divergence, the stabilizing effect of large batch sizes, and cross-entropy-based Lagrange multiplier updates. Moreover, the cumulative toxicity-diversity scores below 0.5 across all baselines indicate significant opportunities for further improvement in balancing toxicity and diversity in adversarial prompt generation.

\paragraph{Limitations.} A primary limitation of this work is the lack of experiments on larger target LLMs, such as those with billions of parameters, due to computational resource constraints. Our experiments were conducted using GPT-2 as both the red-team and target LLM, which, while effective for demonstrating \RedRFT’s capabilities, may not fully capture the complexities of red teaming state-of-the-art models like GPT-4 or LLaMA. Larger models often exhibit different vulnerabilities and behaviors, and evaluating RFT-based red teaming against them could provide deeper insights into scalability and generalizability. Future work could address this limitation by testing \RedRFT\ on more advanced LLMs.

\clearpage
\printbibliography

\appendix

\section{Related Work on Intrinsic Motivation in Reinforcement Learning}
\label{app-sec: related work}

To better illustrate the role of intrinsic motivation in \RedRFT, we provide a brief introduction to the roadmap of intrinsic motivation in RL.

\paragraph{Intrinsic motivation in RL.} Intrinsic motivation enhances exploration of the RL agent by encouraging the agent to seek novel or informative states without relying solely on extrinsic rewards. These techniques are broadly classified into three categories: knowledge-based, data-based, and competence-based intrinsic motivation \cite{laskin2021urlb}. Each category employs distinct mechanisms to guide exploration, but also faces specific limitations. Below, we describe each category, highlighting its principles, applications, and challenges.

\textit{Knowledge-based intrinsic motivation} incentivizes agents to acquire novel knowledge by exploring states with high uncertainty. Rooted in uncertainty quantification, knowledge-based methods often draw inspiration from the Upper Confidence Bound (UCB) algorithm, adapted for nonlinear Markov Decision Processes (MDPs). Typically, a neural network predicts state dynamics, and the prediction error serves as an intrinsic reward, encouraging exploration of less familiar states \cite{burda2018exploration,burda2018large}. Despite its effectiveness in certain settings, this approach struggles in unsupervised or sparse-reward RL scenarios due to issues such as catastrophic forgetting, where previously learned knowledge is lost, and lack of awareness of latent skills \cite{zhang2021made,liu2021aps}.

\textit{Data-based intrinsic motivation} focuses on the agent’s latest interactions with the environment \cite{hazan2019provably,mutti2021task}. A prominent objective in this category is maximizing state entropy, which encourages uniform coverage of the state space by maximizing the differential entropy of the state distribution induced by the current policy \cite{liu2021behavior}. This approach promotes diverse exploration but becomes computationally infeasible in high-dimensional state spaces, where uniform coverage is challenging \cite{park2023metra}.

\textit{Competence-based intrinsic motivation} serves as a powerful intrinsic motivator from an evolutionary perspective. Competence-based methods typically maximize the mutual information between latent skills and the trajectories they generate, enabling agents to develop diverse and reusable skills \cite{gregor2016variational,sharma2019dynamics,laskin2022cic}. However, the invariance of mutual information to scaling and transformations of input variables can result in static skills with limited state-space coverage, hindering their applicability to downstream tasks \cite{park2022lipschitz}. Recent approaches, such as contrastive learning for skill discovery, aim to improve skill diversity and robustness by leveraging structured representations \cite{eysenbach2022contrastive,yang2023behavior}.

This categorization highlights the diversity of intrinsic motivation strategies in RL, each addressing different aspects of exploration. However, integrating these approaches or combining them with extrinsic rewards remains an active area of research to overcome their respective limitations and enhance overall agent performance~\cite{zheng2024constrained}. In the context of RFT-based red teaming, current curiosity/diversity-driven approaches belong to knowledge-based intrinsic motivation, where the intrinsic bonus for each adversarial prompt is estimated based on the rollout buffer of historical adversarial prompts.

\section{Justification of Diversity Score}
\label{app-sec: diversity score}

To evaluate the diversity of the generated adversarial prompts during the fine-tuning process, we propose a novel diversity score as shown in \Cref{eqn: diversity score}. To justify the design of this diversity score, we can easily see that
\begin{equation}
H(\rho_{\va_T}) = \E_{\va_T\sim\gB} [-\ln(\rho_{\va_T})] = \sum_{\va_T\sim\gB} [r^\text{div}].
\end{equation}
That is, the sum of diversity scores of adversarial prompts in the buffer $\gB$ is exactly the differential entropy of $\rho_{\va_T}$, the distribution of the adversarial prompts. When the embeddings of the adversarial prompts are uniformly distributed in the latent space, this differential entropy reaches its maximum.

Note that this diversity score is slightly different from the policy cover-based intrinsic bonus $r_t^\text{PC}$ introduced by CALM~\cite{zheng2025calm}. In $r_t^\text{PC}$, the 

\section{Estimation of Diversity Score}
\label{app-sec: estimation}

To estimate the diversity score $\rho_{\va_T}$, we need to estimate the distribution density $\rho_{\va_T}$. We choose the $k$-Nearest Neighbor ($k$-NN) estimator to avoid complex modeling. Specifically, for an adversarial prompt $\va_T$, its density in the feature space is estimated as follows
\begin{equation}
\label{app-eqn: rho}
\hat\rho_{\va_T}(\va_T) =\frac{\int_{B(\va_T, \kappa)} \rho_{\va_T}(u) \mathrm{d}u}{\int_{B\left(\va_T, r\right)}\mathrm{d}u}\approx \frac{k}{|\gB|\kappa^d},
\end{equation}
where $B$ is the neighborhood of $\va_T$, $\kappa=\|\phi(\va_T)-\phi^*(\va_T)\|$ is the distance between $\phi(\va_T)\in\R^d$ and its $k$-th nearest neighbor $\phi^*(\va_T)$, $d$ is the dimension of the feature space.
Based on \Cref{eqn: rho}, we can obtain the diversity score
\begin{equation}
r^\text{div} = -\ln(\hat\rho_{\va_T}(\va_T)) \approx \ln \kappa
\end{equation}

\section{Discussion on Updating the Lagrange Multiplier}
\label{app-sec: lagrange}

Here we discuss the design details of the Lagrange Multiplier in \RedRFT. The constrained policy optimization problem in \RedRFT\ is
\begin{equation}
\label{app-eqn: lag1}
\max_{\pi'_\alpha} J_\text{tox} + \lambda^\text{I} J_\text{I},\ \text{s.t.}\ J_\text{gib} \le \tau_\text{gib}.
\end{equation}
We use $\pi'_\alpha$ to stand for the policy after update and $\pi_\alpha$ the current policy. Based on the Lagrange dual theory,  \Cref{app-eqn: lag1} becomes
\begin{equation}
\label{app-eqn: lag2}
\min_{\lambda^\text{gib}}\max_{\pi'_\alpha} J_\text{tox} + \lambda^\text{I} J_\text{I} + \lambda^\text{gib} (-J_\text{gib} + \tau_\text{gib}).
\end{equation}
After optimizing the inner maximization problem based on PPO, we can obtain the inner minimization problem
\begin{equation}
\min_{\lambda^\text{gib}} \lambda^\text{gib} (-J^{\pi'_\alpha}_\text{gib} + \tau_\text{gib}).
\end{equation}
where $J^{\pi'_\alpha}_\text{gib} = \E_{\vs\sim\pi'_\alpha} [c^\text{gib}]$ is the expected gibberish score of the adversarial prompts generated by the updated policy $\pi'_\alpha$. The gradient for $\lambda^\text{gib}$ is thus $(-J^{\pi'_\alpha}_\text{gib} + \tau_\text{gib})$ and the update rule of $\lambda^\text{gib}$ becomes
\begin{equation}
\label{app-eqn: lambda}
	\lambda^\text{gib} \leftarrow \lambda^\text{gib} - \eta_{\lambda^\text{gib}} (-J^{\pi'_\alpha}_\text{gib} + \tau_\text{gib}),
\end{equation}
where $\eta_{\lambda^\text{gib}}$ is the learning rate for $\lambda^\text{gib}$. We use the latest adversarial prompts sampled by $\pi'_\alpha$ to estimate $J^{\pi'_\alpha}_\text{gib}$. From \Cref{app-eqn: lambda}, we can easily see that when the constraint is satisfied, $\lambda^\text{gib}$ gradually decreases; otherwise, it gradually grows to strengthen the effect of $J_\text{gib}$ in the inner optimization objective. However, we find in experiments that this update rule heavily relies on the learning rate selection. An inappropriate learning rate will cause severe oscillation of $\lambda^\text{gib}$ and make RFT training unstable.

\section{Details of Algorithmic Baselines}
\label{app-sec: algorithmic baselines}

We present the details of five algorithmic baselines implemented in our \RedRFT\ in this section.

\paragraph{RPPO.} Red-team PPO (RPPO)~\cite{perez2022red} utilizes the basic PPO backbone without any intrinsic rewards.

\paragraph{TDiv.} Different from CRT and DiveR-CT that define the intrinsic rewards based on the sentence embeddings of the adversarial prompts, TDiv.~\cite{casper2023explore} defines the intrinsic rewards based on the sentence embeddings of the target responses induced by the adversarial prompts, that is, "the intra-batch cosine distances of the target LM’s embeddings of the generated prompts." Since the original paper does not provide a formal definition of this reward, we define this intrinsic reward as follows:
\begin{equation}
\label{app-eqn: rew-tdiv}
r_t^\text{TDiv} = \begin{cases}
0 & \text{if } t\in\{0, ..., T-2\} \\ 
\sum_{\vy'\sim\pi_\alpha} -\phi(\vy) \phi(\vy') & \text{if } t = T-1
\end{cases},
\end{equation}
where $\vy$ is the target response as defined in \Cref{eqn: rollout}. Note that we use $\vy'\sim\pi_\alpha$ instead of $\vy'\sim\gB$ to obey the description of the original paper, that is, "intra-batch cosine distances". $r_t^\text{TDiv}$ is also a prompt-level intrinsic reward, that is, only when $t=T-1$, this reward can be non-zero.

\paragraph{CRT.} In CRT~\cite{hong2024curiosity}, apart from the prompt-level intrinsic reward $r_t^\text{Cos}$, Hong et al. also propose another prompt-level intrinsic reward based on SelfBLEU~\cite{zhu2018texygen}. The BLEU score assesses how similar two sentences are, while the SelfBLEU score evaluates how one sentence resembles the rest in a generated collection. Regarding one sentence as a hypothesis and the others as references, we can calculate the BLEU score for every generated sentence.
\begin{equation}
r_t^\text{BLEU} = \begin{cases}
0 & \text{if } t\in\{0, ..., T-2\} \\ 
1 - \text{BLEU}(\va_T,\gB) & \text{if } t = T-1
\end{cases},
\end{equation}
where $\text{BLEU}(\va_T,\gB)$ calcuates the BLEU score of the hypothesis sentence $\va_T$ based on the refenrece sentences from $\gB$. Though the definition is simple, we find in experiments that the calculation of the BLEU score becomes too time-consuming when the rollout buffer $\gB$ grows. Hence, to balance the accuracy and efficiency, we only sample $\min(3\times\text{batch\_size},|\gB|)$ adversarial prompts from the rollout buffer $\gB$ to calculate the $r_t^\text{BLEU}$. The composite intrinsic reward of CRT is 
\begin{equation}
r_t^\text{CRT} = r_t^\text{Cos} + r_t^\text{BLEU}
\end{equation}
Since CRT does not employ the constrained policy optimization, the final optimization objective for CRT is
\begin{equation}
\max_{\pi_\alpha} J_\text{tox} + \lambda^\text{CRT} J_\text{CRT} - \lambda^\text{gib}J_\text{gib}
\end{equation}
where $J_\text{CRT} = \E[r_t^\text{CRT}]$ is the intrinsic objective of CRT, $\lambda^\text{CRT}$ is the coefficient to balance the extrinsic objective $J_\text{tox}$ and the intrinsic objective.

\paragraph{DiveR-CT.} DiveR-CT uses a similar prompt-level intrinsic reward as CRT. The difference lies in the sampling strategy. Instead of utilizing the whole samples from the rollout buffer, DiveR-CT proposes to use only the $k$-th nearest neighbors of the adversarial prompts to estimate the novelty, that is,
\begin{equation}
r_t^\text{kCos} = \begin{cases}
0 & \text{if } t\in\{0, ..., T-2\} \\
\sum_{\va'_T\sim B(\va_T)} -\phi(\va_T) \phi(\va'_T) & \text{if } t = T-1
\end{cases},
\end{equation}
where $B(\va'_T)$ indicates the $k$-th nearest neighbors of the adversarial prompt $\va_T$. The composite intrinsic reward of DiveR-CT is then
\begin{equation}
r_t^\text{DiveR-CT} = r_t^\text{kCos} + r_t^\text{BLEU}.
\end{equation}
DiveR-CT adopts the constrained policy optimization, and the final optimization objective is
\begin{equation}
\label{app-eqn: diverct}
\max_{\pi_\alpha} J_\text{tox} + \lambda^\text{DiveR-CT} J_\text{DiveR-CT},\ \text{s.t.}\ J_\text{gib} \le \tau_\text{gib},
\end{equation}
where $J_\text{DiveR-CT} = \E[r_t^\text{DiveR-CT}]$ is the intrinsic objective of CRT, $\lambda^\text{DiveR-CT}$ is the coefficient to balance the extrinsic objective $J_\text{tox}$ and the intrinsic objective. Note that in the original paper of DiveR-CT, the authors also investigate the cases where there exists a constraint for $J_\text{tox}$. However, to standardize the RFT-based red teaming, we choose to set the default optimization objective of DiveR-CT as \Cref{app-eqn: diverct}. It is easy and flexible to configure the constraint for any objective, e.g.,  $J_\text{tox}$, in our \RedRFT\ (with only one line change in the configuration file).

\paragraph{CALM.} CALM~\cite{zheng2025calm} proposes a state-level intrinsic reward based on the policy cover theory as defined in \Cref{eqn: rew-pc}. We adopt a constrained policy optimization objective for CALM as follows:
\begin{equation}
\max_{\pi_\alpha} J_\text{tox} + \lambda^\text{CALM} J_\text{CALM},\ \text{s.t.}\ J_\text{gib} \le \tau_\text{gib},
\end{equation}
where $J_\text{CALM} = \E[r_t^\text{PC}]$ is the intrinsic objective of CALM and $\lambda^\text{CALM}$ is the corresponding coefficient. Note that in the original paper of CALM, the authors leverage the technique of Random Network Distillation (RND)~\cite{burda2018exploration} to estimate the policy cover and state density. To make the comparison fair, we adopt the non-parametric $k$-NN estimator in \Cref{app-eqn: rho} instead of RND to estimate the policy cover and state density.

\section{Implementation Details}
\label{app-sec: implementation}

\paragraph{Models and datasets.} For both tasks, we use GPT-2 as the red-team model. When fine-tuning the red-team model, we either fine-tune only the last two blocks or utilize the Low-Rank Adaptation (LoRA) technique.
For text continuation, as mentioned in \Cref{sec: framework}, we use IMDB as the dataset to sample an initial prompt for the red-team LLM. The target model in text continuation is also GPT-2. For instruction following, we adopt Alpaca as the dataset and GPT2-Alpaca\footnote{\url{https://huggingface.co/vicgalle/gpt2-alpaca}} as the target model. The running time of one experiment varies from 30 minutes to 1 hour.

\paragraph{Hyperparameters.} We list all hyperparameters for the optimization backbone in \Cref{app-tab: ppo-hyper} and five algorithmic baselines in \Cref{app-tab: rew-hyper}.

\begin{table}[t]
\centering
\caption{Hyperparameters for the PPO Backbone in \RedRFT}
\label{app-tab: ppo-hyper}
\begin{tabular}{>{\raggedright\arraybackslash}p{5cm}>{\raggedright\arraybackslash}p{5cm}}
\toprule
\textbf{Parameter} & \textbf{Value} \\
\midrule
Total Queries & 50,000 \\
Batch Size & 256 \\
Learning Rate & $3 \times 10^{-5}$ \\
Mini-Batch Size & 16 \\
Gamma & 0.99 \\
Epochs & 4 \\
Normalize Returns & True \\
Value Function Coefficient & 1 \\
Policy Entropy Coefficient & 0.001 \\
KL Reference Coefficient & 0.001 \\
Maximum Gradient Norm & 1 \\
Lagrange Type & Average \\
Lagrange Learning Rate & 0.1 \\
\bottomrule
\end{tabular}
\end{table}

\begin{table}[t]
\centering
\caption{Reward Coefficients and Constraints for Algorithmic Baselines in Reinforcement Fine-Tuning-Based Red Teaming}
\label{app-tab: rew-hyper}
\begin{tabular}{>{\raggedright\arraybackslash}p{3cm}>{\raggedright\arraybackslash}p{5cm}>{\raggedright\arraybackslash}p{4cm}}
\toprule
\textbf{Method} & \textbf{Reward Coefficients} & \textbf{Cost Threshold} \\
\midrule
EX & $\lambda^\text{tox}=1,\lambda^\text{gib}=-1$ & -- \\
TDiv & $\lambda^\text{TDiv}=1,\lambda^\text{gib}=-1$ & -- \\
CRT & $\lambda^\text{CRT}=1,\lambda^\text{gib}=-1$ & -- \\
DiverCT & $\lambda^\text{TDiv}=1$ & $\tau_\text{gib}=0.1$ \\
CALM & $\lambda^\text{CALM}=0.1$ & $\tau_\text{gib}=0.1$ \\
RedRFT & $\lambda^\text{RedRFT}=1$ & $\tau_\text{gib}=0.1$ \\
\bottomrule
\end{tabular}
\end{table}

\section{Ablation Study on Intrinsic Reward}
\label{app-sec: ablation on intrinsic rew}

\begin{figure}
\centering
\includegraphics[width=\linewidth]{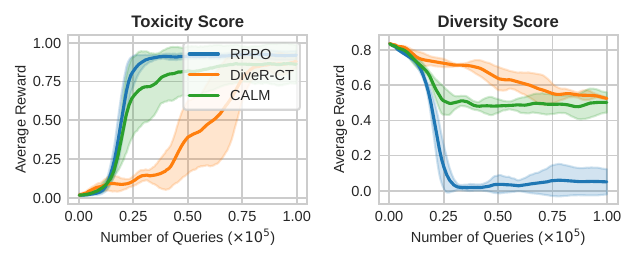}
\caption{Learning curves of toxic rewards and intrinsic rewards during the fine-tuning process.}
\label{app-fig: calm_vs_diverct}
\end{figure}

As mentioned in \Cref{sec: experiments}, state-level intrinsic rewards produce results that are comparable to those of prompt-level intrinsic rewards. To illustrate this key finding more clearly, we visualize the variations in toxic rewards and intrinsic rewards during the fine-tuning process in \Cref{app-fig: calm_vs_diverct}. The results indicate that CALM can identify high-quality adversarial prompts more efficiently than DiveR-CT, albeit at the cost of reduced diversity in the adversarial prompts. Furthermore, when compared to RPPO, which does not incorporate any intrinsic rewards, both CALM and DiveR-CT demonstrate a better balance between toxicity scores and diversity scores.

\section{Ablation Study on Batch Size}
\label{app-sec: ablation on batch size}

\begin{figure}
\centering
\includegraphics[width=0.3\linewidth]{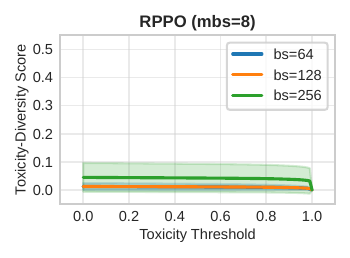}
\includegraphics[width=0.3\linewidth]{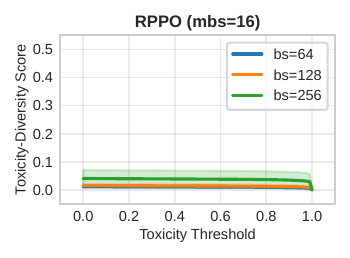}
\includegraphics[width=0.3\linewidth]{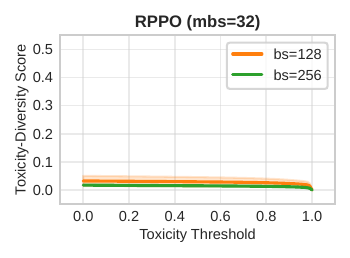}
\includegraphics[width=0.3\linewidth]{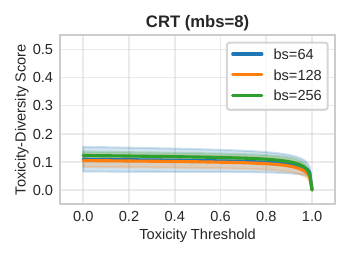}
\includegraphics[width=0.3\linewidth]{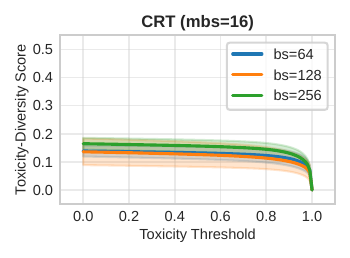}
\includegraphics[width=0.3\linewidth]{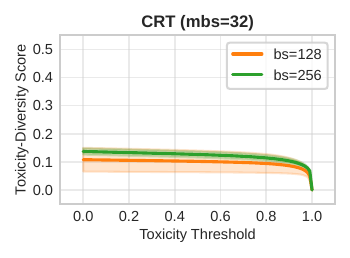}
\includegraphics[width=0.3\linewidth]{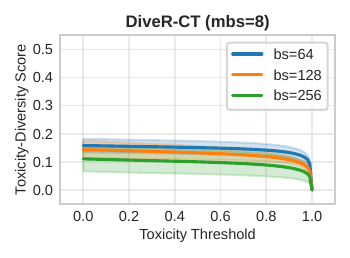}
\includegraphics[width=0.3\linewidth]{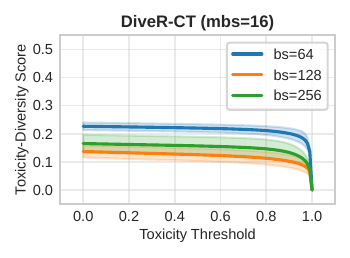}
\includegraphics[width=0.3\linewidth]{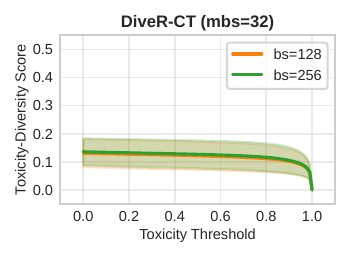}
\caption{Ablation study on batch size in toxic completion. bs means batch\_size and mbs stands for mini\_batch\_size. We find that batch\_size=256 and mini\_batch\_size=16 show the most comparable results across different methods.}
\label{app-fig: full_bs_tc}
\end{figure}

\begin{figure}
\centering
\includegraphics[width=0.3\linewidth]{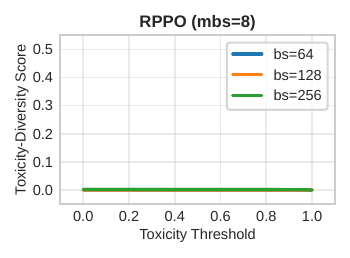}
\includegraphics[width=0.3\linewidth]{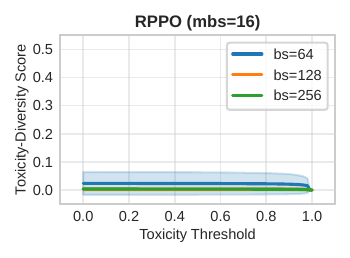}
\includegraphics[width=0.3\linewidth]{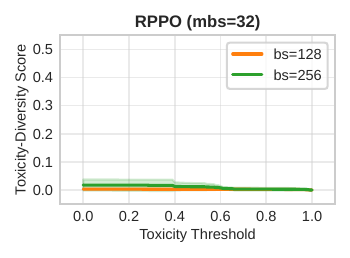}
\includegraphics[width=0.3\linewidth]{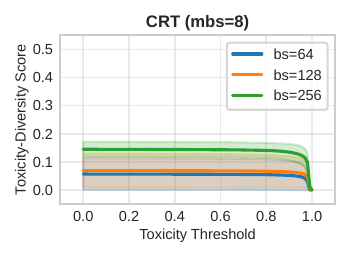}
\includegraphics[width=0.3\linewidth]{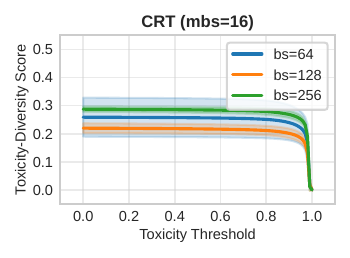}
\includegraphics[width=0.3\linewidth]{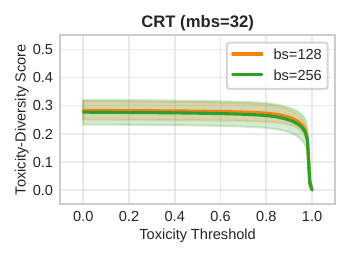}
\includegraphics[width=0.3\linewidth]{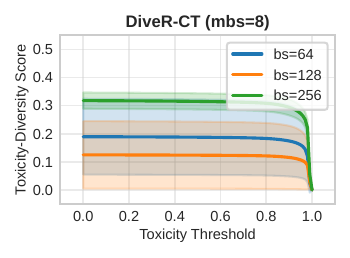}
\includegraphics[width=0.3\linewidth]{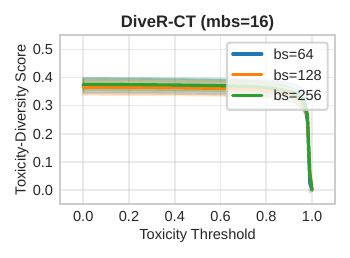}
\includegraphics[width=0.3\linewidth]{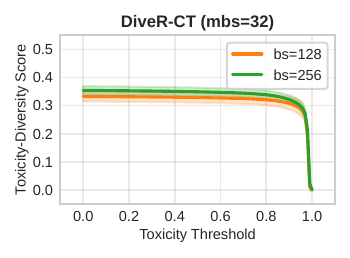}
\caption{Ablation study on batch size in instruction following. bs means batch\_size and mbs stands for mini\_batch\_size. We find that batch\_size=256 and mini\_batch\_size=16 show the most comparable results across different methods.}
\label{app-fig: full_bs_if}
\end{figure}

We provide the full results of the ablation study on batch size across three methods, RPPO, CRT, and DiveR-CT, in both toxic completion and instruction following, shown in \Cref{app-fig: full_bs_tc} and \Cref{app-fig: full_bs_if}. We choose batch size from $[64,128,256]$ and mini-batch size from $[8,16,32]$. We exclude the config of batch\_size=64 and mini\_batch\_size=32 since the training under such a config is unstable. From the results, we can conclude that a large batch size is more suitable for RFT-based red teaming for better toxicity-diversity score and efficiency.

\section{Ablation Study on LoRA and KL Divergence}
\label{app-sec: ablation on lora}

\begin{figure}
\centering
\includegraphics[width=\linewidth]{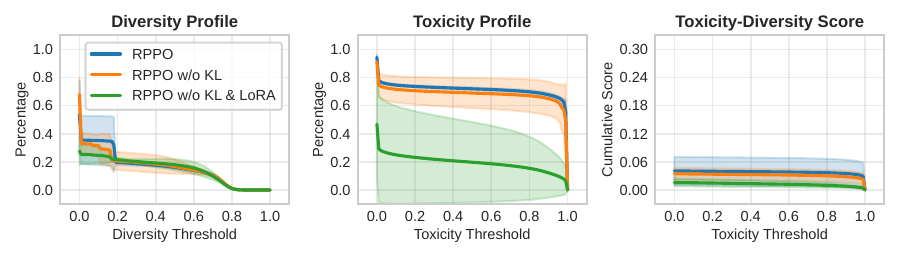}
\includegraphics[width=\linewidth]{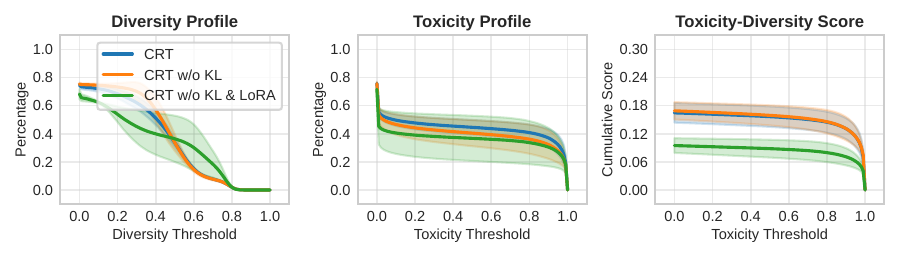}
\includegraphics[width=\linewidth]{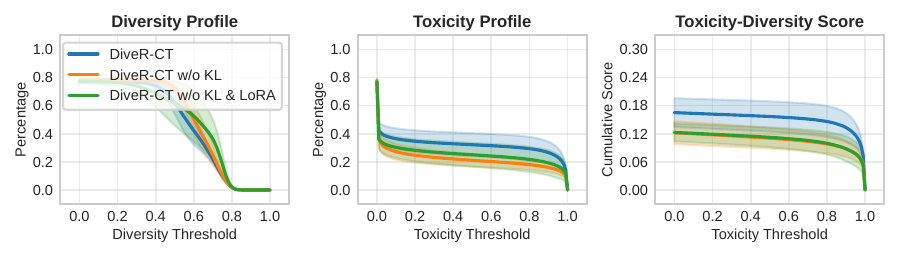}
\caption{Ablation study on LoRA and KL divergence in toxic completion. LoRA and KL divergence both contribute to the performance across different algorithmic baselines.}
\label{app-fig: full_lora_tc}
\end{figure}

We present detailed results from our ablation study on LoRA and KL divergence. As illustrated in \Cref{app-fig: full_lora_tc}, both LoRA and KL divergence enhance performance across various algorithmic baselines.

\section{Ablation Study on Lagrangian Multiplier}
\label{app-sec: ablation on Lag}

\begin{figure}
\centering
\includegraphics[width=\linewidth]{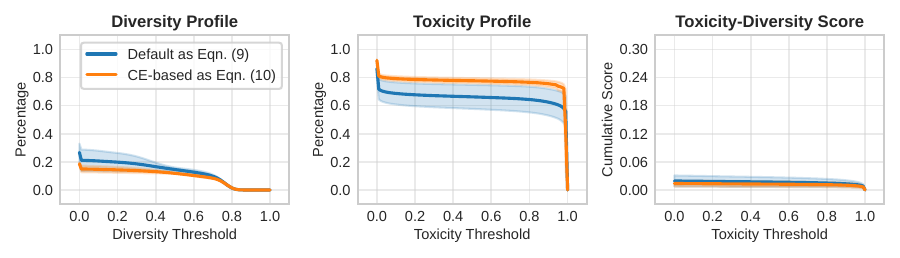}
\includegraphics[width=\linewidth]{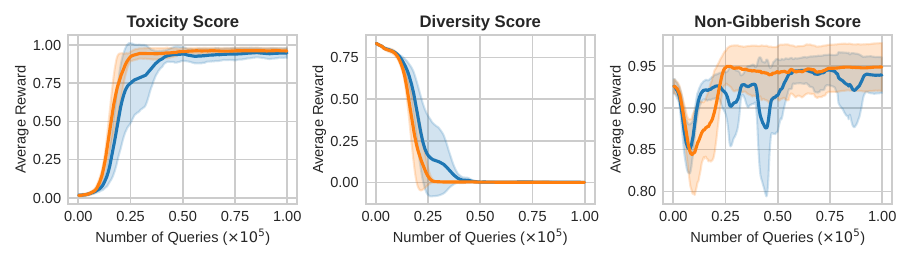}
\caption{Ablation study on Lagrangian Multiplier}
\label{app-fig: full_lag_tc}
\end{figure}

For the update of Lagrangian multiplier, apart from the learning curve of the Lagrangian multiplier presented in the main body of this paper, we also provide a detailed result on the corresponding toxic scores and non-gibberish scores. The results shown in \Cref{app-fig: full_lag_tc} underscore the importance of the stable update rule for the Lagrangian multiplier.

\section{Demonstration of Fast Prototyping}
\label{app-sec: fastproto}

\begin{figure}
\centering
\includegraphics[width=\linewidth]{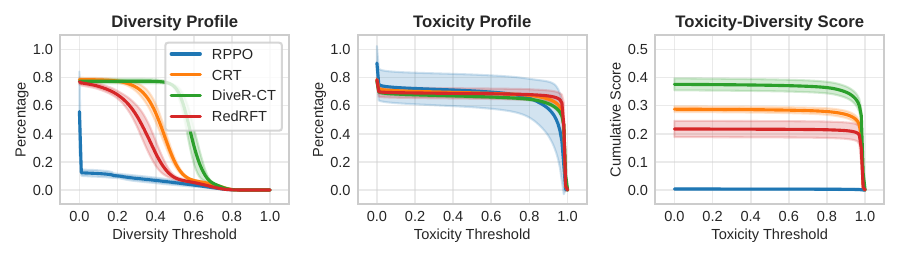}
\caption{RedRFT demonstrates comparable performance to CRT and DiveR-CT in instruction following as a prototype method.}
\label{app-fig: full_redrft_if}
\end{figure}

To further illustrate the advantages of \RedRFT\ in fast prototyping, we present the results of a novel RFT-based red teaming method, RedRFT. This method modifies only the intrinsic reward of DiveR-CT, changing it to the entropy-based diversity score \( r^{\text{div}} \). As shown in Figure \ref{app-fig: full_redrft_if}, RedRFT exhibits performance that is comparable to other state-of-the-art RFT-based red teaming methods in the instruction-following task.

\section{Potential Societal and Ethical Impact}
\label{app-sec: impact}

As \RedRFT\ focuses on RFT-based red teaming, our experiments generate adversarial prompts that elicit toxic or unsafe responses from target LLMs. While this process raises ethical concerns due to the potential creation of harmful content, we emphasize that these experiments are conducted in a controlled research environment with the explicit goal of improving LLM safety. By identifying vulnerabilities in LLMs, \RedRFT\ will facilitate the development of more robust safety alignment techniques, which are critical for mitigating risks associated with real-world deployment. The toxic content generated during our experiments is not disseminated but used solely to evaluate and enhance the resilience of LLMs against adversarial prompts. Furthermore, \RedRFT’s standardized framework promotes transparency and reproducibility, enabling the research community to build on our findings responsibly. We believe that the societal benefits of safer and more trustworthy LLMs outweigh the controlled risks of our experiments, as these efforts contribute to reducing the potential for harm in applications ranging from conversational agents to automated content generation. Nonetheless, we advocate for ongoing ethical oversight and adherence to responsible AI research practices to ensure that red teaming advancements align with societal values.

\newpage
\section*{NeurIPS Paper Checklist}

\begin{enumerate}

\item {\bf Claims}
    \item[] Question: Do the main claims made in the abstract and introduction accurately reflect the paper's contributions and scope?
    \item[] Answer: \answerYes{} 
    \item[] Justification: Please refer to the main body of the paper, especially \Cref{sec: framework}, \Cref{sec: algorithmic baselines}, \Cref{sec: experiments}.

\item {\bf Limitations}
    \item[] Question: Does the paper discuss the limitations of the work performed by the authors?
    \item[] Answer: \answerYes{} 
    \item[] Justification: Please refer to \Cref{sec: conclusion}.

\item {\bf Theory assumptions and proofs}
    \item[] Question: For each theoretical result, does the paper provide the full set of assumptions and a complete (and correct) proof?
    \item[] Answer: \answerNA{} 
    \item[] Justification: Our benchmark RedRFT focuses on the framework standardization of the RFT-based red teaming and empirical ablation study on implementation details. Thus, we do not provide original theoretical results of the each baseline method. For strict theoretical results of these baseline methods, please refer to their original papers.

    \item {\bf Experimental result reproducibility}
    \item[] Question: Does the paper fully disclose all the information needed to reproduce the main experimental results of the paper to the extent that it affects the main claims and/or conclusions of the paper (regardless of whether the code and data are provided or not)?
    \item[] Answer: \answerYes{} 
    \item[] Justification: The design principle of our benchmark RedRFT is light-weight and clean, as described in \Cref{sec: framework} and \Cref{sec: algorithmic baselines}. Our experiment results can be easily reproduced with a simple environment setup, the default config files, and the prepared shell scripts. Please refer to \Cref{sec: experiments} and \Cref{app-sec: implementation}.

\item {\bf Open access to data and code}
    \item[] Question: Does the paper provide open access to the data and code, with sufficient instructions to faithfully reproduce the main experimental results, as described in supplemental material?
    \item[] Answer: \answerYes{} 
    \item[] Justification: Please refer to \url{https://github.com/x-zheng16/RedRFT.git} for the code of our benchmark.

\item {\bf Experimental setting/details}
    \item[] Question: Does the paper specify all the training and test details (e.g., data splits, hyperparameters, how they were chosen, type of optimizer, etc.) necessary to understand the results?
    \item[] Answer: \answerYes{} 
    \item[] Justification: Please refer to \Cref{sec: experiments} and \Cref{app-sec: implementation}.

\item {\bf Experiment statistical significance}
    \item[] Question: Does the paper report error bars suitably and correctly defined or other appropriate information about the statistical significance of the experiments?
    \item[] Answer: \answerYes{} 
    \item[] Justification: We report error bars in all our experiments, as shown in \Cref{sec: experiments}.

\item {\bf Experiments compute resources}
    \item[] Question: For each experiment, does the paper provide sufficient information on the computer resources (type of compute workers, memory, time of execution) needed to reproduce the experiments?
    \item[] Answer: \answerYes{} 
    \item[] Justification: Please refer to \Cref{sec: experiments}.
    
\item {\bf Code of ethics}
    \item[] Question: Does the research conducted in the paper conform, in every respect, with the NeurIPS Code of Ethics \url{https://neurips.cc/public/EthicsGuidelines}?
    \item[] Answer: \answerYes{} 
    \item[] Justification: We confirm that our Benchmark conforms, in every respect, with the NeurIPS Code of Ethics.

\item {\bf Broader impacts}
    \item[] Question: Does the paper discuss both potential positive societal impacts and negative societal impacts of the work performed?
    \item[] Answer: \answerYes{} 
    \item[] Justification: Since the main paper has limited space, we include the discussion on the potential societal and ethical impact of our benchmark in \Cref{app-sec: impact}.
    
\item {\bf Safeguards}
    \item[] Question: Does the paper describe safeguards that have been put in place for responsible release of data or models that have a high risk for misuse (e.g., pretrained language models, image generators, or scraped datasets)?
    \item[] Answer: \answerNA{} 
    \item[] Justification: The baseline methods our benchmark reproduces are all publicly published, and there is no high risk of misuse in these methods, as discussed in their original papers. 

\item {\bf Licenses for existing assets}
    \item[] Question: Are the creators or original owners of assets (e.g., code, data, models), used in the paper, properly credited and are the license and terms of use explicitly mentioned and properly respected?
    \item[] Answer: \answerYes{} 
    \item[] Justification: We respect and cite all the main Python libraries our benchmark relies on, including TensorDict, Hydra, and PyKeops.

\item {\bf New assets}
    \item[] Question: Are new assets introduced in the paper well documented and is the documentation provided alongside the assets?
    \item[] Answer: \answerYes{} 
    \item[] Justification: We provide a detailed document for all new assets. Please refer to \url{https://github.com/x-zheng16/RedRFT.git} for the document.

\item {\bf Crowdsourcing and research with human subjects}
    \item[] Question: For crowdsourcing experiments and research with human subjects, does the paper include the full text of instructions given to participants and screenshots, if applicable, as well as details about compensation (if any)? 
    \item[] Answer: \answerNA{} 
    \item[] Justification: Our benchmark belongs to automated red teaming and does not involve crowdsourcing or research with human subjects.

\item {\bf Institutional review board (IRB) approvals or equivalent for research with human subjects}
    \item[] Question: Does the paper describe potential risks incurred by study participants, whether such risks were disclosed to the subjects, and whether Institutional Review Board (IRB) approvals (or an equivalent approval/review based on the requirements of your country or institution) were obtained?
    \item[] Answer: \answerNA{} 
    \item[] Justification: Our benchmark belongs to automated red teaming and does not involve crowdsourcing or research with human subjects.

\item {\bf Declaration of LLM usage}
    \item[] Question: Does the paper describe the usage of LLMs if it is an important, original, or non-standard component of the core methods in this research? Note that if the LLM is used only for writing, editing, or formatting purposes and does not impact the core methodology, scientific rigorousness, or originality of the research, declaration is not required.
    \item[] Answer: \answerNA{} 
    \item[] Justification: The core design and development process of our benchmarks, including the standardized RFT-based red teaming framework and the implementation of all RFT-based red teaming methods, does not rely on LLMs. We only use LLMs, e.g., GPT-4o, for grammar correction.

\end{enumerate}

\end{document}